\documentclass{article}

\usepackage{arxiv}

\usepackage[utf8]{inputenc} % allow utf-8 input
\usepackage[T1]{fontenc}    % use 8-bit T1 fonts
\usepackage{hyperref}       % hyperlinks
\usepackage{url}            % simple URL typesetting
\usepackage{booktabs}       % professional-quality tables
\usepackage{amsfonts}       % blackboard math symbols
\usepackage{nicefrac}       % compact symbols for 1/2, etc.
\usepackage{microtype}      % microtypography
\usepackage{graphicx}
\usepackage{natbib}
\usepackage{doi}
\usepackage{authblk}
\usepackage{chngcntr}
\usepackage{makecell}
\usepackage{longtable}
\usepackage{subcaption}

\title{Zero Shot Health Trajectory Prediction Using Transformer}

\author[1,2,3]{Pawel Renc}
\author[4]{Yugang Jia}
\author[1,2]{Anthony E. Samir}
\author[3]{Jaroslaw Was}
\author[1,2]{Quanzheng Li}
\author[5,2,6]{David W. Bates}
\author[1,2,*]{Arkadiusz Sitek}

\affil[1]{Massachusetts General Hospital, Boston, USA}
\affil[2]{Harvard Medical School, Boston, USA}
\affil[3]{AGH University of Science and Technology, Krakow, PL}
\affil[4]{Massachusetts Institute of Technology, Cambridge, USA}
\affil[5]{Brigham and Women’s Hospital, Boston, USA}
\affil[6]{Harvard Chan School of Public Health, Boston, USA}
\affil[*]{\textit{Corresponding author: sarkadiu@gmail.com}}
\date{}

\renewcommand{\headeright}{}
\renewcommand{\undertitle}{preprint}
\renewcommand{\shorttitle}{Zero Shot Health Trajectory Prediction Using Transformer}

\newcommand\SupplementaryMaterials{%
  \xdef\presupfigures{\arabic{figure}}% save the current figure number
  \xdef\pretables{\arabic{table}}% save the current section number
  \renewcommand\thefigure{S\fpeval{\arabic{figure}-\presupfigures}}
  \renewcommand\thetable{S\fpeval{\arabic{table}-\pretables}}
}

\begin{document}
\maketitle

\begin{abstract}
Integrating modern machine learning and clinical decision-making has great promise for mitigating healthcare's increasing cost and complexity. We introduce the Enhanced Transformer for Health Outcome Simulation (ETHOS), a novel application of the transformer deep-learning architecture for analyzing high-dimensional, heterogeneous, and episodic health data. ETHOS is trained using  Patient Health Timelines (PHTs)—detailed, tokenized records of health events—to predict future health trajectories, leveraging a zero-shot learning approach. ETHOS represents a significant advancement in foundation model development for healthcare analytics, eliminating the need for labeled data and model fine-tuning. Its ability to simulate various treatment pathways and consider patient-specific factors positions ETHOS as a tool for care optimization and addressing biases in healthcare delivery. Future developments will expand ETHOS’ capabilities to incorporate a wider range of data types and data sources. Our work demonstrates a pathway toward accelerated AI development and deployment in healthcare.
\end{abstract}

% keywords can be removed

\section{Introduction}
Healthcare in the U.S. is the world’s most expensive, and the quality and safety of care do not compare well to other developed countries~\cite{schneider_mirror_2021}. While electronic healthcare records are now ubiquitous in the U.S., and decision-support technologies are widely implemented, most are rule-based, and their effectiveness so far has been limited~\cite{bates_improving_2022}. Artificial intelligence has emerged as a technique with great potential for improving care, but most organizations are not using it to any major degree. Two major limiting factors have been (1)  the lack of large, labeled datasets, which are expensive and time-consuming to develop; and (2) limited system capacity to deliver recommendations to the appropriate clinician at the optimal time.  In this manuscript, we describe a novel method called the Enhanced Transformer for Health Outcome Simulation (ETHOS), which we believe can help address many of the limitations that have prevented widespread AI adoption.

\begin{figure}
    \centering
    \includegraphics[width=1\linewidth]{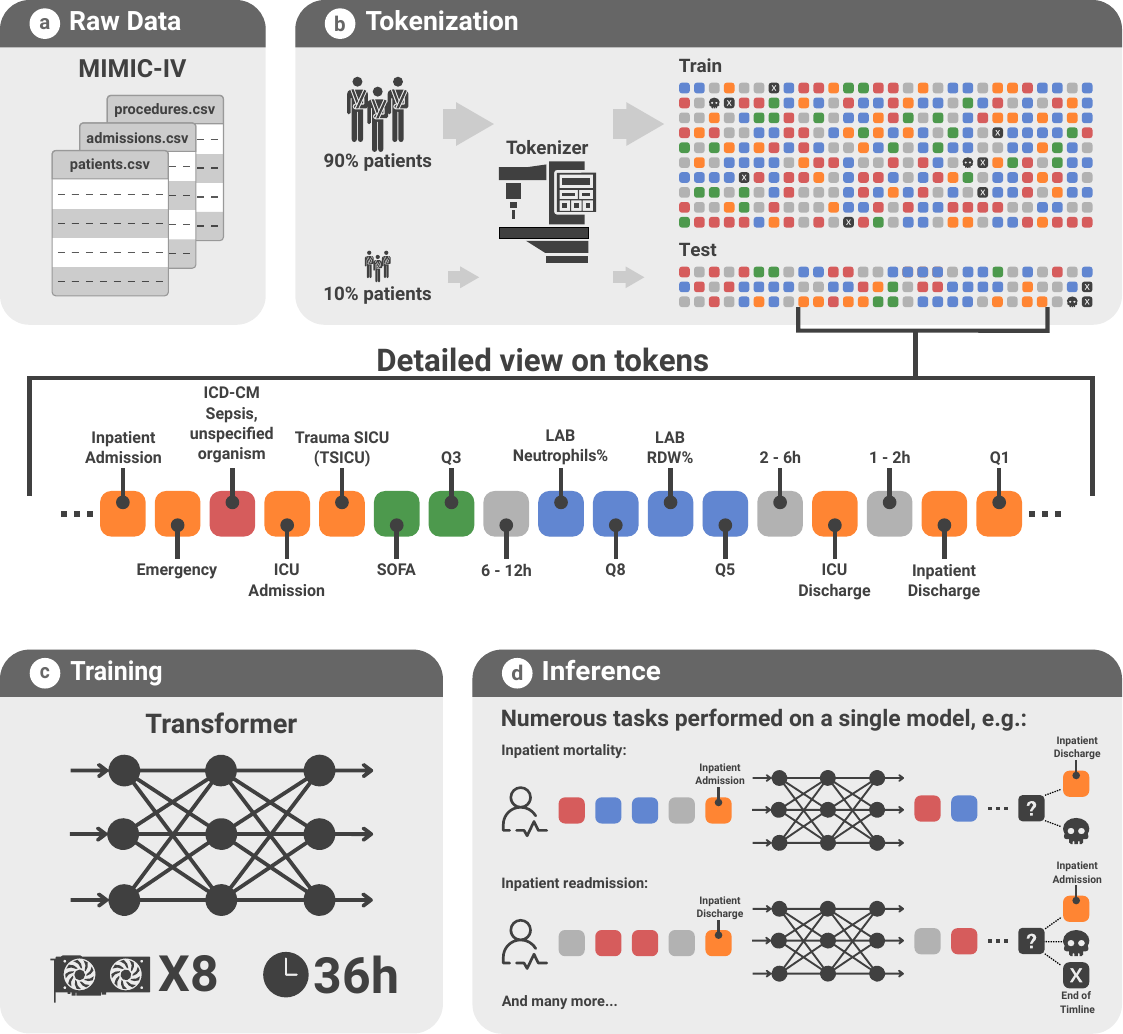}
    \caption{
        \textbf{Implementing the ETHOS Model with EMR Data.}
        (a) Extraction of raw patient data from the MIMIC-IV database, encompassing tables of admissions, patient demographics, medical procedures, among others.
        (b) The tokenization process, utilizing data from 90\% of patients for model training and the remaining 10\% for testing, transforms complex medical records into structured PHT for efficient model processing.
        (c) Training phase illustration, employing a transformer architecture optimized across 8 GPUs over a span of 36 hours.
        (d) Demonstration of ETHOS's zero-shot inference capabilities, highlighting its proficiency in performing tasks such as predicting inpatient mortality and readmission rates, leveraging forecasted future PHTs.
    }
    \label{fig:pipeline}
\end{figure}

ETHOS is a novel application of the transformer deep-learning architecture, originally conceptualized for natural language processing~\cite{vaswani_attention_2017}. This architecture, a cornerstone in large language model (LLM) development, is repurposed in ETHOS to analyze health-related data, moving beyond the textual focus of traditional LLMs. ETHOS is designed to process Patient Health Timelines (PHTs)—detailed tokenized chronological records of health-related events—to predict future health timelines. In PHTs, a token serves as the fundamental unit of information, encapsulating diverse data types such as patient admissions, administered medications, or time intervals. We elaborate on this pivotal aspect of our methodology in the Methods section. Our model takes the patient's health history, as represented by PHT, and subsequently forecasts future PHT (fPHT) on a token-by-token basis (refer to Figure~\ref{fig:pipeline}).

ETHOS’s generative capabilities are gained in unsupervised learning. Once trained, ETHOS can forecast future health events without requiring task-specific training. This is done through a zero-shot learning approach, making ETHOS a versatile foundation model for numerous healthcare applications. With appropriate modifications, ETHOS can be adapted to a broad range of data types, including but not limited to medical images, clinical and discharge notes, monitoring data, data from wearables, or omics data.

In this research, we leverage the recently released MIMIC-IV~v.2.2 dataset~\cite{johnson_mimic-iv_2023}, a rich open-source repository accompanied by our code, allowing others to replicate our findings. MIMIC-IV is expansive, chronicling more than 400,000 hospitalizations in more than 200,000 patients. Although relatively large, we anticipate that the performance of our system will further improve as we expand the dataset with additional patient histories and data types. 

Importantly, we utilize the MIMIC-IV dataset in its original noisy form without any data modifications, cleaning, or targeted imputation for missing entries. The information is retained in the face of large data inconsistencies, such as discharge dates noted before admission dates. We operated under the assumption that, within large enough datasets and appropriate tokenization and training methods, ETHOS would be robust enough to handle the noisy input and automatically manage the noise/anomalies in the input data. The resilience of ETHOS to data inaccuracies and missing information has important implications for the efficiency of downstream model development. Healthcare data inevitably contains errors, some of which may not be immediately apparent or easily rectifiable. Attempts to clean large datasets can be impractical and may inadvertently introduce biases and errors. Our approach highlights the vital need for algorithms adept at managing these challenges, a prerequisite for the large-scale development of reliable and robust healthcare AI applications. 

Our research showcases the zero-shot learning capabilities of ETHOS in predicting inpatient and ICU mortality, estimating ICU length of stay (LOS), and determining readmission probabilities. Additionally, we illustrate the model's versatility by performing a regression task to estimate the first-day Sequential Organ Failure Assessment (SOFA) score~\cite{raith_prognostic_2017} at the time of ICU admission using information before admission (see example in Figure~\ref{fig:pipeline}d). The SOFA score is a critical tool for monitoring a patient's condition in the ICU, evaluating organ function or failure across six systems—respiratory, cardiovascular, hepatic, coagulation, renal, and neurological—with each system scored from 0 to 4, culminating in a total possible minimum score of 0 and maximum score of 24. Furthermore, we predict Diagnostic-Related Group (DRG) classifications, encompassing over 771 categories, at the time of hospital discharge. The DRG system categorizes hospital cases into standardized case complexity-based Medicare and Medicaid payment groups, encouraging efficient patient care without compromising quality. The diversity of tasks ETHOS can perform, from mortality predictions and LOS estimation to SOFA scoring and DRG classification, highlights its broad applicability and zero-shot learning efficiency. 

ETHOS is a foundation model~\cite{moor_foundation_2023}, introducing a novel approach in the landscape of data analysis within the healthcare domain. The other foundational models developed recently have fallen into two broad categories. The first of these categories encompasses Clinical Language Models (CLaMs), a specialized subset of large language models (LLMs)~\cite{brown_language_2020} tailored for processing clinical and biomedical text data. These models are typically trained on extensive datasets containing clinical notes, biomedical literature, and other healthcare-related text sources. CLaMs are proficient in various clinical tasks such as extracting drug names, summarizing medical dialogues, predicting clinical outcomes, and responding to patient queries~\cite{wornow_shaky_2023,zack_assessing_2024,li_fine-tuning_2019,jiang_health_2023,wang_drg-llama_2024}. The second category comprises Foundation Models for Electronic Medical Records (FEMRs), representing another class of clinical foundation models tailored specifically for EMR data analysis. FEMRs undergo training on the extensive medical histories of patients, covering both structured data (such as demographics and lab results) and unstructured data (including progress notes and radiology reports). Unlike CLaMs, FEMRs are not designed to generate clinical text. Instead, they produce machine-understandable representations of patient data, facilitating tasks such as patient phenotyping and outcome prediction~\cite{jiang_health_2023, steinberg_language_2021, li_hi-behrt_2023, savcisens_using_2024}. Similarly, data that chronicles human lives, akin to EMR, can also be modeled effectively in this manner. 

The primary distinction between ETHOS and previously published methods lies in our approach, which eliminates the need for fine-tuning or labeled data to produce accurate inferences or predictions. We demonstrate inference across a wide array of tasks without task-specific training. Moreover, the ability of ETHOS to forecast future PHTs opens the door to a wide array of bespoke and innovative applications, facilitating its use in unique scenarios in healthcare, some of them explored in the discussion section. Unlike many studies, which often apply specific criteria for selecting data for training and testing, our methodology imposes no such limitations. This feature is crucial for considering the scalability of the ETHOS approach to data sets comprising millions or even hundreds of millions of patients.

\section{Results}

\subsection{Tokenization of MIMIC data and training of ETHOS}

\begin{figure}
    \centering
    \includegraphics[width=1\linewidth]{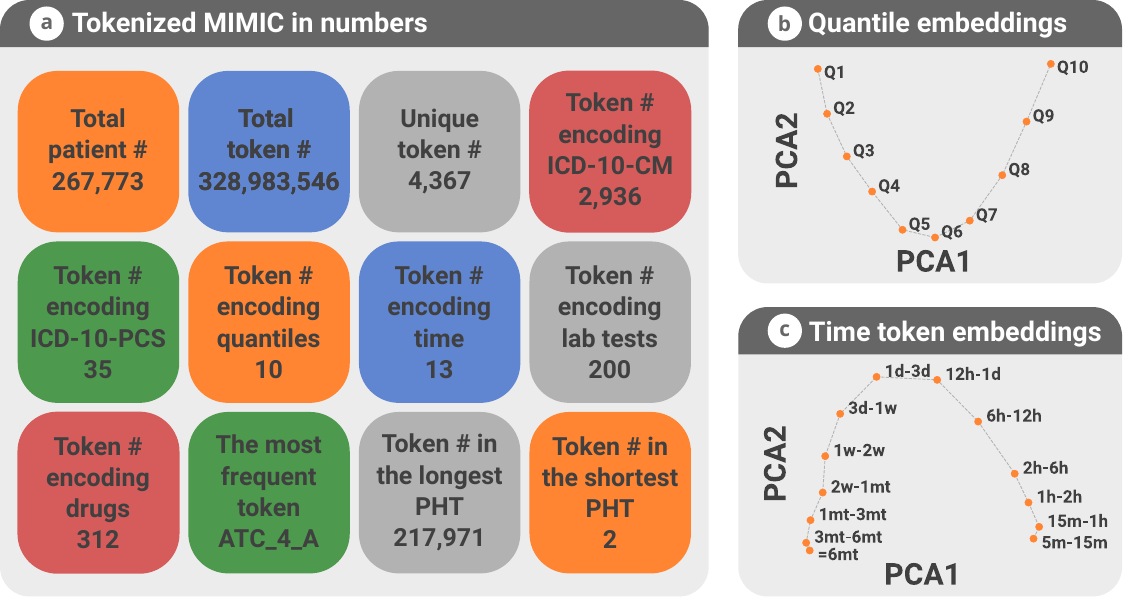}
    \caption{
        \textbf{Tokenization and Embedding Visualizations of MIMIC-IV Data.}
        (a) Overview of key insights derived from the tokenization process applied to MIMIC-IV data.
        (b) Visualization of embedding vectors for quantile tokens (Qs), which categorize quantitative information across the dataset. Each quantitative measure (e.g., blood pressure) is encoded by a preceding category-specific token followed by a quantile token, delineating its position within a predefined value range. This method facilitates a structured, scalable representation of complex data types via a systematic token sequence.
        (c) Visualization of embedding vectors for time-interval tokens, illustrating the temporal distribution and relationships within the PHT.
    }
    \label{fig:token-summary}
\end{figure}

\begin{figure}[ht]
    \centering
    \begin{subfigure}[b]{0.45\linewidth}
        \centering
        \includegraphics[width=\textwidth]{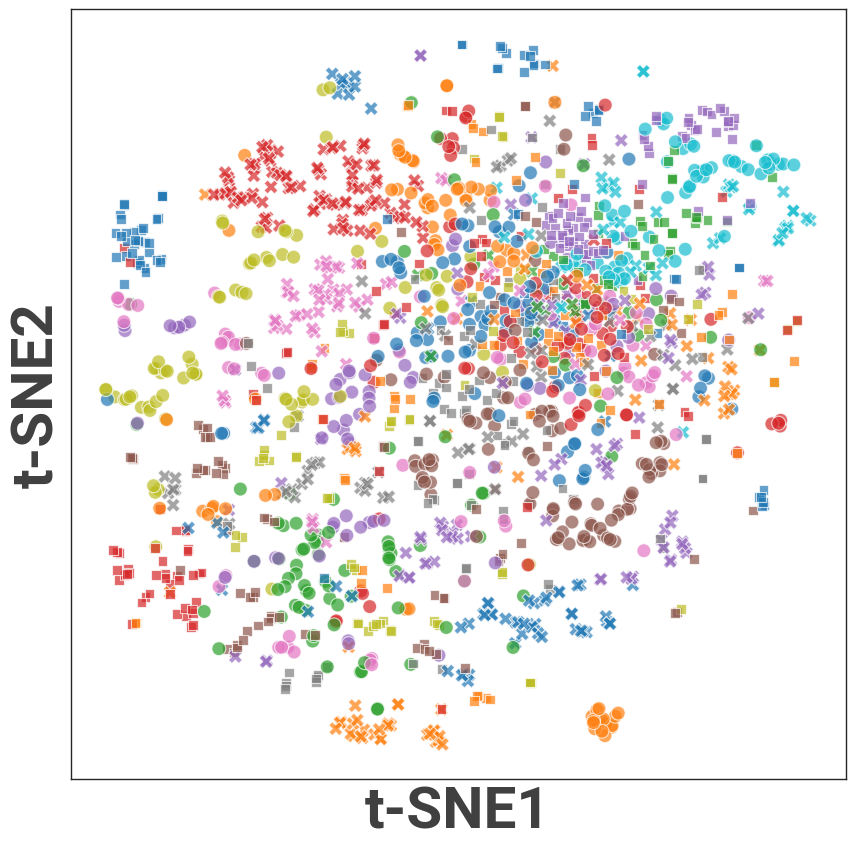}
        \caption{t-SNE graph of embeddings of tokens to represent three first characters of ICD10-CM code.  Points in the graph are differentiated by color and symbol type representing the first letter of an ICD code (25 categories) which broadly indicate diagnostic area. Note that these broad diagnostic areas only partially are separated.}
    \end{subfigure}
    \hfill
    \begin{subfigure}[b]{0.45\linewidth}
        \centering
        \includegraphics[width=\textwidth]{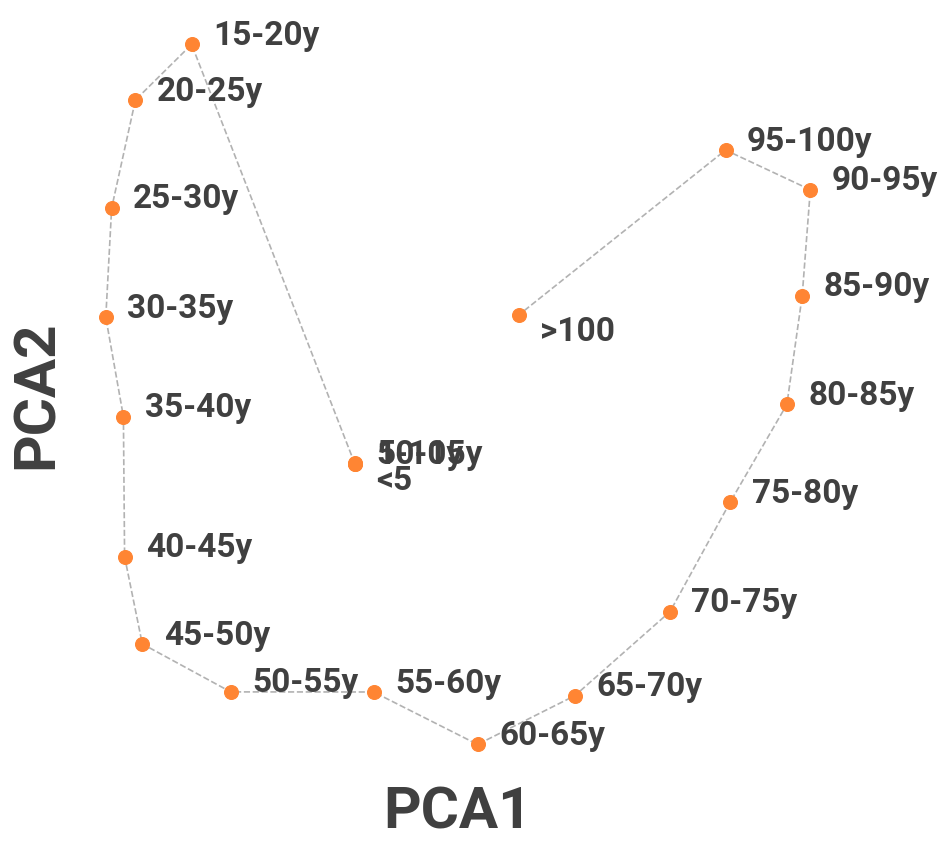}
        \caption{Visualization of embedding vectors for quantile tokens representing patient age and used to also indicate start of the PHT. This is a similar visualization as in the main paper Figure 2 for quantiles and time-interval tokens.  As before we discern a certain structure in which tokens are correctly ordered in PCA reduced 2D embedding space.}
    \end{subfigure}
    \caption{\textbf{Learned token embeddings projected into 2 dimensions.}}
    \label{supfig:3}
\end{figure}

Figure~\ref{fig:token-summary}a summarizes some statistics of the tokenization process, including the number of tokens generated and other details. Figure~\ref{supfig:3}b presents visualizations of the 768-dimensional embeddings reduced to a 2D plane using Principal Component Analysis (PCA) for quantile tokens, which encode all quantitative values in the data. The tokens are arranged from Q1 (the lowest quantile) to Q10 (the highest quantile). This suggests that the transformer model has learned a sequential relationship between the tokens that mirrors their natural order, ascertaining this order from the data during the training process. The proximity between points could reflect the model's differentiation among the quantiles. We observe that the gaps between Q4, Q5, and Q6 are narrower than those between Q9 and Q10. This may suggest that the model deems the variance between population-average values to be less substantial than that of extremely high values. For example, the difference in clinical significance between a blood pressure reading of 110 mmHg (Q5) and one of 130 mmHg (Q6) is less pronounced than the difference between 140 mmHg (Q9) and 160 mmHg (Q10), which could account for the greater disparity in the embedding vectors of high quantiles.

The embeddings for time-interval tokens, representing the approximate durations between different tokenized events in PHT, are illustrated in Figure~\ref{supfig:3}b. These embeddings display a pattern analogous to that observed for Q tokens, where ETHOS systematically arranged them according to the actual time values they represent. Remarkably, the model perceives the two shortest (5m-15m, 15m-1h), and two longest (3m-6m, 6m) intervals as relatively similar. 

\subsection{ETHOS inferences}

\begin{figure}
    \centering
    \includegraphics[width=1\linewidth]{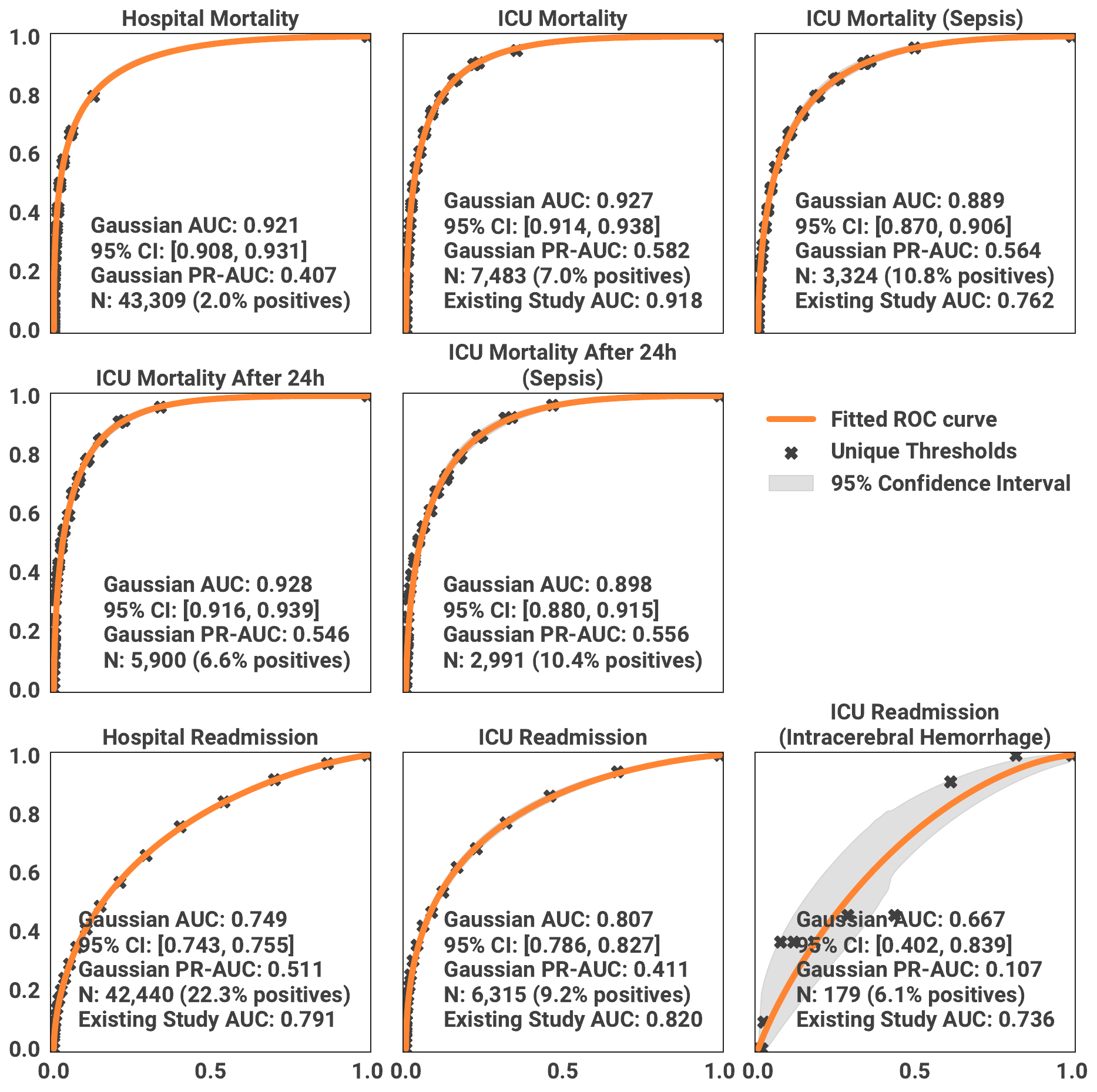}
    \caption{
        \textbf{Receiver Operating Characteristic (ROC) Curves for Predictive Tasks via the ETHOS Model.}
        Each graph delineates the model's efficacy in forecasting distinct clinical outcomes, specifically mortality and readmission rates. Accompanying each ROC curve are the case count (N), the outcome prevalence, and the 95\% confidence interval for the AUC. Points marked with an 'X' denote specific thresholds utilized for classification decisions within the ETHOS model. Area under precision-recall (PR) curves is also provided and PR-curves are presented in supplementary material. The AUC of the existing study represents the performance of the best algorithms identified in the literature, with references provided within the text.
    }
    \label{fig:results}
\end{figure}

\begin{figure}[ht]
    \centering
    \includegraphics[width=1\linewidth]{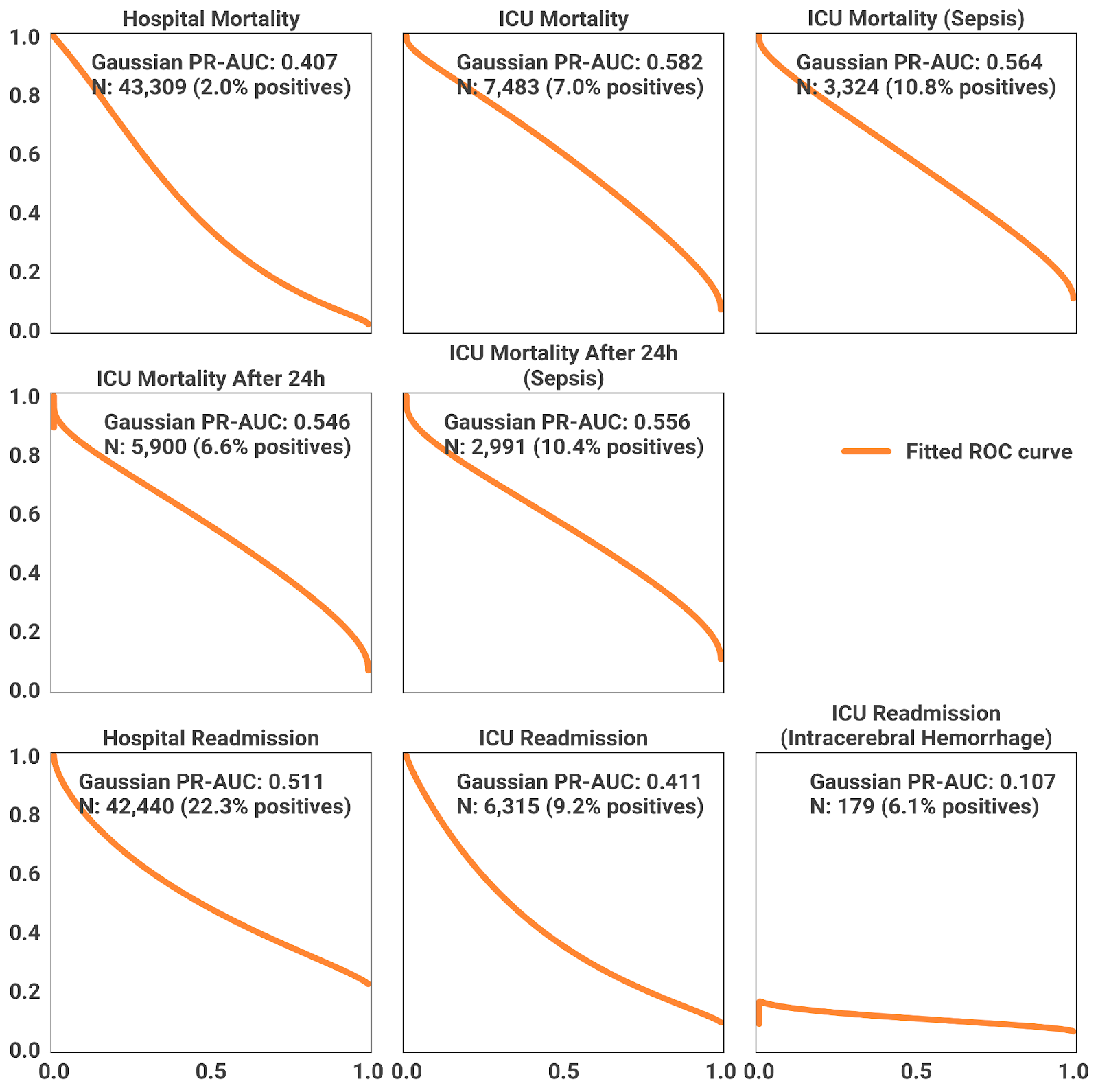}
    \caption{
        \textbf{Precision-recall curves and corresponding scores for Predictive Tasks via the ETHOS Model.}
    }
    \label{fig:sup-7}
\end{figure}

In our study, we conducted zero-shot inferences for a diverse array of classification tasks, including readmission to the ICU, inpatient mortality, ICU mortality, combined inpatient and ICU mortality in patients with sepsis, readmission to the ICU for patients with intracerebral hemorrhage, assignment of DRG class assessed at inpatient discharge. We also demonstrate regression of first-day SOFA score at the time of ICU admission and regression of the length of stay in ICU in days assessed upon admission. The results corresponding to these tasks are summarized in Figure~\ref{fig:results}. We also provide precision-recall curves of the corresponding results in Figure~\ref{fig:sup-7}.

To situate our results within the broader scientific discourse, we conducted a literature review, concentrating on contemporary studies that utilized the MIMIC-III and MIMIC-IV datasets for similar tasks and reported their outcomes. A notable observation from our review is that many of these studies either lacked publicly available source code or implemented specific exclusion criteria for their data selection. Such practices pose challenges for directly comparing their results with our approach. Nonetheless, we posit that the numerical outcomes reported in these works provide a valuable benchmark for assessing the performance of ETHOS. Furthermore, we conducted a direct comparative analysis of ETHOS against specialized algorithms developed in-house, with these findings detailed in the supplementary materials.

We conducted an analysis focusing on risk estimation for inpatient and ICU mortality, calculated at the respective points of patient admission to the hospital and ICU. The test set comprised 43,309 hospital admissions with a 2.0\% mortality and 7,483 ICU admissions with a 7.0\% mortality. The ETHOS model demonstrated robust performance, achieving an AUC of 0.921 (95\% CI: 0.908-0.931) for hospital mortality and 0.927 (95\% CI: 0.914-0.938) for ICU mortality. Comparatively, in the ICU mortality risk prediction domain, the highest performance identified in our literature review was an AUC of 0.918 (95\% CI: 0.915-0.922) reported by~\cite{pang_establishment_2022} using the XGBoost model. On the lower end, \cite{chen_deep_2023} reported an AUC of 0.642 ± 0.101. Within a specific subgroup of the test set of 3,324 patients with sepsis with 10.8\% mortality prevalence, ETHOS's prediction of ICU mortality exhibited an AUC of 0.889 (95\% CI: 0.870-0.906), which is a better performance than obtained in a study by~\cite{pan_evaluate_2023}, which estimated ICU mortality in adult sepsis patients using SOFA and additional features, achieving an AUC of 0.762 ± 0.006. We also estimated performance for a task of ICU mortality estimation for patients remaining in ICU for at least 24 hours in which we obtained an AUC of 0.928 (95\% CI: 0.916-0.939). 

Furthermore, ETHOS estimated the length of stay (LoS) in the ICU with a mean absolute error (MAE) of 2.262 days (95\% CI: 2.161-2.355 days). These results paralleled those of18, who reported an MAE of 2.42 ± 0.10 days. ICU LoS prediction and mortality risk, underscoring the competitive zero-shot performance of ETHOS across multiple key healthcare metrics.

For the ICU readmission task, ETHOS’ AUC of 0.807 (95\% CI: 0.786-0.827) is slightly smaller than the AUC of 0.82 obtained using knowledge graph embeddings~\cite{carvalho_knowledge_2023} and is higher than the AUC of 0.791 (95\% CI, 0.782–0.800) using LSTMs based on MIMIC-III data~\cite{lin_analysis_2019}. Additionally, we applied our method to a task characterized by a relatively low prevalence, specifically focusing on only 174 cases of patients with hemorrhage admitted to the ICU present within our test set. The prediction of readmission by ETHOS yielded an AUC of 0.667 (95\% CI: 0.402-0.839), comparable to the AUC of 0.736 (95\% CI: 0.668-0.801) achieved by previous studies~\cite{miao_predicting_2024} using LightGBM. For hospital readmission, ETHOS achieved an AUC of 0.749 (95\% CI: 0.743-0.755), lower than the AUC of  0.791 [0.766-0.816] obtained by~\cite{tang_predicting_2023}. It's important to recognize that although MIMIC offers a wealth of data on acute care, it might not encompass all the subtleties necessary for readmission research, including comprehensive post-discharge outcomes or data on readmissions to various hospitals. Consequently, the accuracy of results for tasks related to readmission may be limited, regardless of the method employed.

\begin{figure}
    \centering
    \includegraphics[width=1\linewidth]{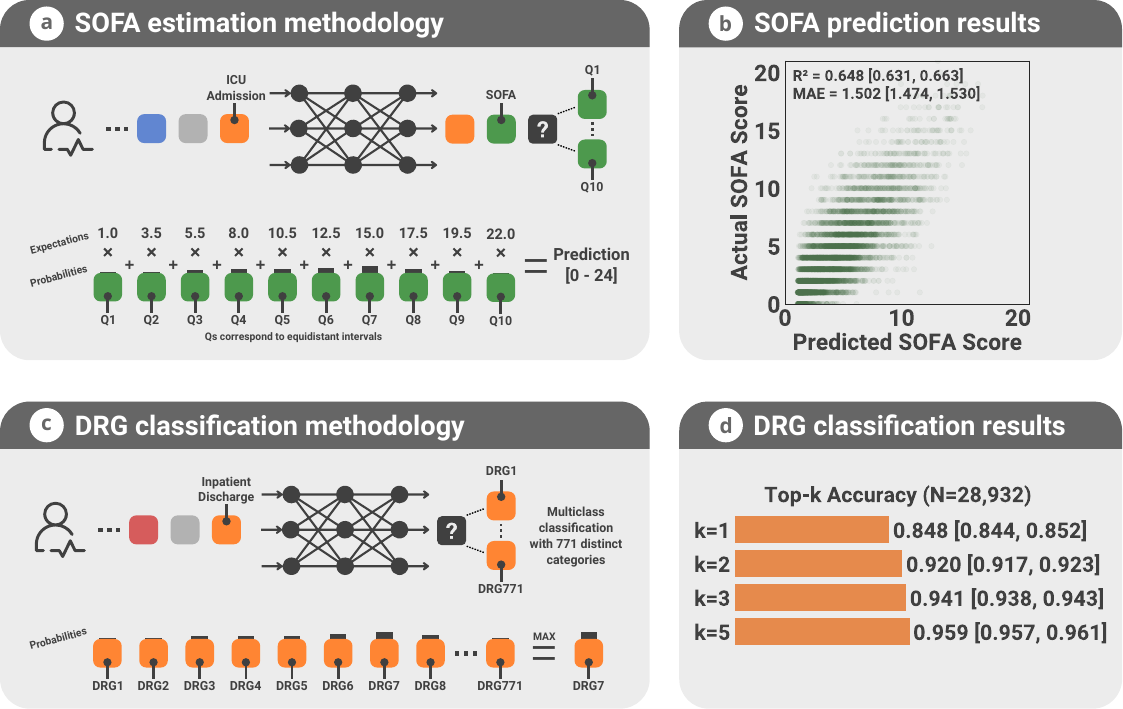}
    \caption{
        \textbf{ETHOS Model Performance on SOFA Estimation and DRG Classification.}
        (a) Estimation of the first-day Sequential Organ Failure Assessment (SOFA) score at ICU admission by ETHOS, which generates a sequence of three tokens: the admission type (orange token), a SOFA token (indicating the SOFA score estimation will follow), and a quantile token (q-token indicated by question mark) predicting probabilities of the SOFA score's quantile, as detailed at the bottom of the panel (a). The fixed position of the SOFA token ensures its consistent prediction immediately after ICU admission. The SOFA score is derived using quantile probabilities generated by ETHOS and average value of SOFA for ten quantiles (values of 1.0, 3.5 …). Since SOFA value 24 was not present in the dataset we predict values 0-23.
        (b) Correlation plot between actual and predicted SOFA scores. 
        (c) For Diagnostic Related Groups (DRG) classification. The model is trained to insert a DRG token after tokens typically used at discharge time, utilizing a placeholder “DRG\_UNKNOWN” for if DRG is unknown in the training set. Predicted probabilities are used to compute the top-{1,2,3,5} DRG classifications.
        (d) Visualization of DRG classification accuracy, showcasing the model's predictive performance.
    }
    \label{fig:results-sofa-drg}
\end{figure}

We explored the task of predicting the first-day SOFA score at the time of admission (Figure~\ref{fig:results-sofa-drg}). Given that the SOFA score is a critical indicator of survival, particularly in sepsis~\cite{raith_prognostic_2017,minne_evaluation_2008}, this prediction can serve as a valuable indirect prognostic marker of ICU patient health status. We achieved a SOFA score estimation with an MAE of 1.502 (95\% CI: 1.475-1.534). To our knowledge, no prior literature predicts first-day SOFA at the time of admission.

For the DRG assignment, we observed a top-1 (out of 771 classes) accuracy rate of 84.8\% (95\% CI: 84.4\%-85.2\%) in 28,932 hospitalizations using our methodology, a significant improvement over the 52\% reported by~\cite{wang_drg-llama_2024}, who explored DRG estimation using LLMs from discharge notes. This marked enhancement in performance can be attributed to the comprehensive nature of ETHOS, which incorporates a wide array of clinical events leading up to discharge within the PHT. In contrast, the approach taken by~\cite{wang_drg-llama_2024} relies solely on discharge notes, which may not encompass the breadth of information captured by PHT, thus potentially explaining the disparity in accuracy rates. 

We want to reiterate an important point: all comparisons presented in this section are made between ETHOS, trained indiscriminately on the entire test population and task-specific algorithms developed using much smaller MIMIC data subsets obtained after data curation. In addition to the results in this section, in supplementary materials, we benchmark the performance of ETHOS against XGBoost~\cite{chen_xgboost_2016}, recurrent neural networks, and logistic regression. 

\section{Discussion}

This work introduces an innovative approach to developing a Foundation Model for medical data derived from EMRs, designed to execute zero-shot inferences across a diverse range of tasks. Our model generates interpretable, causally forecasted future patient health timelines. By "causal," we mean that predictions are made solely based on information that occurred in the past. We applied and evaluated this model using the MIMIC-IV EMR datasets, comparing its performance with the results of methods published in the literature for the same tasks. Our objective, however, was not merely to surpass the performance of these specialized SOTA implementations. Instead, we aimed to demonstrate that ETHOS, a single foundation model trained just once with zero-shot derived inference, can achieve performance levels comparable to that of multiple models optimized for various tasks. This underscores the potential of ETHOS to streamline the application of AI in healthcare by leveraging a single unified model development architecture and set of methods for multiple prediction tasks, thereby greatly enhancing medical data model development efficiency and scalability.

The application of patient timelines for generating insights has been established in existing research~\cite{jiang_health_2023,steinberg_language_2021, li_hi-behrt_2023, savcisens_using_2024, bornet_comparing_2023}, as has the implementation of foundation models~\cite{moor_foundation_2023}. Our methodology sets itself apart by integrating a zero-shot capability, obviating the need for additional training beyond the initial model. The design of ETHOS accommodates various approaches to inference, including few-shot predictions, although this necessitates fine-tuning for specific downstream tasks. Notably, the zero-shot prediction methodology introduces capabilities absent in few-shot prediction. To forecast future outcomes, ETHOS generates multiple health timelines representing possible future scenarios. This functionality exploits the model's capacity to explore and evaluate potential future events, thereby potentially estimating uncertainties. Future work will undoubtedly concentrate on refining this aspect of ETHOS. Moreover, ETHOS is specifically engineered to produce causal predictions in the form of future timelines, ensuring they are inherently comprehensible to human users. This is achieved through a novel tokenization process for medical data, a distinctive feature of our work.

The generation of multiple scenarios using the zero-shot approach places significant demands on time and computational resources. We estimated the inference time based on the average duration required to generate 1,000 tokens. On a single Nvidia A100 GPU, this process took approximately 15 seconds. Given that computations are executed in batches and are thus highly parallelizable, we anticipate that in a potential production environment, response times could vary between 1 to 30 seconds. This variation is contingent upon the complexity of the downstream task at hand.

Another highly distinctive capability of ETHOS is the potential to generate individualized care-integrated PHT-based projected healthcare expenditures. This capability is exemplified through the prediction of Diagnosis-Related Group (DRG) codes but is not limited to this application. Specifically, ETHOS can model future PHTs at critical decision-making junctures in patient care. For instance, ETHOS can model outcomes for administering either drug A or B, considering the patient's unique conditions (such as sex, age, race, gender, income, etc.) to determine which path might yield better clinical and cost outcomes. In this regard, ETHOS has the potential to revolutionize medical decision-analytic modeling science by incorporating a level of personalization previously unavailable in conventional decision-analytic models. This has the potential to enhance clinical decision-making and incorporate individualized real-time quantitatively robust value-based care policies into clinical care.  This is a potentially transformative change, radically  unlike current evidence-based medicine practices, which  rely on high-quality data obtained from and averaged across patient populations~\cite{zack_assessing_2024, obermeyer_dissecting_2019, abid_large_2021}.

In designing ETHOS, we have considered explainability, fairness, and transparency. These are vital aspects of our ongoing research. In future work, we plan to implement and test advanced visualization attention layers of the transformer~\cite{vig_multiscale_2019} to gain insights into the model's reasoning process. Additionally, a dedicated interface for decision-making is envisaged further to enhance the usability of ETHOS in clinical settings.

Envisioning the development of a robust AI method that offers fully personalized advice on a wide range of medical questions necessitates learning from an extensive dataset of patients. Such a model must assimilate as much data as possible and be adaptable to a vast array of medical tasks. ETHOS represents a significant stride in this direction. Built on a transformer architecture, it is inherently scalable and, as a zero-shot learner, is versatile enough to address numerous key medical prediction tasks without task-specific training. Currently, ETHOS does not incorporate various types of critical information, including clinical and discharge notes, medical imaging and pathology images, genetic data, socioeconomic factors, lifestyle considerations, and monitoring signals. Nonetheless, the conceptual framework for incorporating these diverse data types is relatively straightforward. This can be done by leveraging the encoder and cross-attention mechanisms inherent in the transformer architecture; we anticipate the potential for integrating a nearly limitless amount of information during training. This expansion of ETHOS's capabilities forms the cornerstone of our future work, promising to enhance its applicability and efficacy in personalized medical advice and diagnostics.

We aim to modify further and train ETHOS to apply it across diverse data sources. This capability is currently hindered by variations in data collection methodologies, disparities in data quality, and the presence or absence of certain data types across different sources. Additionally, non-overlapping populations present significant challenges, rendering ETHOS not yet generalizable. To mitigate some of these compatibility issues, we propose the development of a universal tokenization format~\cite{mcdermott_event_2023}. While this approach may resolve certain discrepancies, it does not address all underlying compatibility concerns. The ultimate solution, we believe, lies in a system capable of transforming tokenized data from one healthcare system to another, akin to text translation between languages. Specifically, for ETHOS, this would mean converting the patient journey, as encapsulated by the PHT, from one system's format to another. This conversion would not only facilitate a consistent and unified representation of patient histories across different systems but also offer insights into the operational nuances of these systems. Pursuing such a translation strategy represents a vital direction for our future research endeavors, alongside evaluating the methodologies introduced in this paper through analysis of prospectively collected data.

ETHOS and LLMs such as GPT-4o, Claude 3 Opus, and Gemini 1.5 Ultra, although built upon similar AI principles, serve different purposes and exhibit distinct capabilities. ETHOS is specifically designed to predict fPHTs through explicit modeling of quantitative values and temporal sequences. This approach allows ETHOS to leverage structured patient data to generate predictions. In contrast, LLMs are general-purpose models optimized for tasks involving knowledge integration, reasoning, and interactive conversation. They do not explicitly model quantitative values and time sequences, which are important for accurate clinical decision support. Studies, such as those by~\cite{hager_evaluation_2024} and~\cite{wang_tram_2024}, highlight the limitations of LLMs in handling temporal information and decision support tasks, emphasizing the potential need for specialized models like ETHOS. There is a potential of ETHOS to be used in conjunction with LLMs through retrieval-augmented generation (RAG) mechanisms, offering a promising direction for future AI applications in healthcare. In supplementary material,  we present a comparison in predictive performance of ETHOS and LLM (GPT-4o). Furthermore, while LLMs excel in processing vast amounts of unstructured text, their computational performance in generating detailed and contextually accurate patient predictions remains suboptimal compared to ETHOS because of efficient representation of information in ETHOS tokenized PHTs.

This work has limitations.  We utilized the MIMIC dataset, which may be cleaner than many routine clinical datasets. Performance and usability should be tested prospectively in diverse datasets and in real-time. The transformer model in the current version of ETHOS is relatively simple and uses only 2048 PHT tokens for predictions. When token density per time is large, this may not contain sufficient information for optimal performance. Mitigation of the limitation is expected with additional computational infrastructure.

In conclusion, ETHOS presents a promising approach to deriving insights from massive clinical datasets without labor-intensive labeling or distinct model creation for each prediction task. This approach has the potential to significantly lower the costs and complexities associated with AI model development, thereby accelerating the development and implementation of healthcare AI.

\section{Methods}

\subsection{Data}

In this study, the Medical Information Mart for Intensive Care (MIMIC-IV) database served as a data source, providing a rich and comprehensive collection of de-identified health-related information4. Managed collaboratively by the Massachusetts Institute of Technology (MIT), Beth Israel Deaconess Medical Center (BIDMC), and Philips Healthcare, MIMIC-IV encompasses detailed records for more than 200,000 patients who were admitted to hospital and critical care units at BIDMC in Boston, Massachusetts, between 2008 and 2019. The following tables from the MIMIC-IV were used: 1) Patients, which contains static information about the patients, such as gender, date of birth, and date of death; 2) Admissions, which holds information about patient admissions to the hospital, including admission and discharge times, as well as information related to the hospital stay; 3) Icustays, which is specifically related to intensive care unit (ICU) stays, including the timings and type of ICU; 4) Labevents, which contains laboratory test results for patients. We used the 200 most frequent tests covering 95\% of tests completed; 5) Prescriptions, which holds information on medications prescribed to patients during their stay, with each drug converted to ATC code~\footnote{\url{https://www.whocc.no}}. We converted GSN codes in MIMIC-IV to ATC codes using conversion tables~\cite{bornet_comparing_2023}; 6) Procedures which contains information about procedures performed on patients, coded using ICD10-PCS codes; 7) Diagnoses which contains diagnostic information, typically coded using ICD10-CM codes. We converted ICD9 to ICD10-CM if needed using conversion table~\footnote{\url{https://www.cms.gov/medicare/coding-billing/icd-10-codes}}; 8) Emar, which holds information related to the documentation and administration of medications to patients; 9) Omr with information about measurements taken from a patient, such as blood pressure or BMI; 10) Services with information about the clinical service under which a patient is managed during their hospital stay; 11) drgcodes DRG codes which are a classification system used in the healthcare industry to categorize hospital cases into groups that are expected to have similar hospital resource use;  12) SOFA, taken from the derived tables in MIMIC. The remaining tables were not used in the current ETHOS implementation as they will require additional processing. For example, clinical notes require natural language processing to be converted to meaningful tokenized information.

\begin{figure}
    \centering
    \includegraphics[width=1\linewidth]{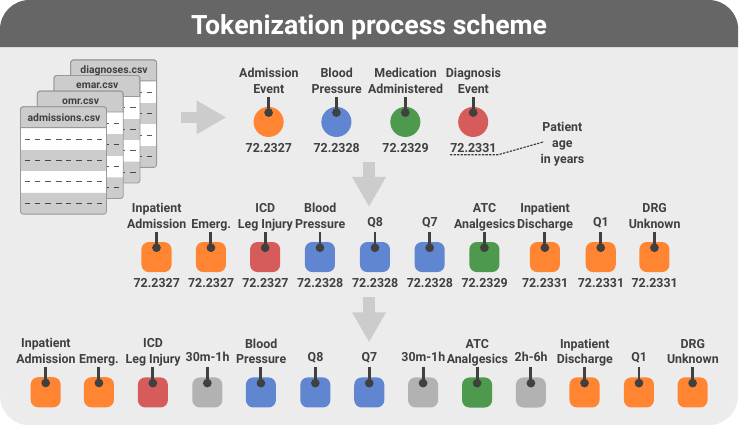}
    \caption{
        \textbf{Stages of PHT Construction and Tokenization in ETHOS}
        The process begins with assembling a chronological list of events from MIMIC-IV tables, Each entry on the list is time stamped with 64-bit real value only 6 significant digits show for clarity, indicating the patient's age at which the event occurred. Subsequently, list elements are transformed into tokens using ETHOS tokenization scheme. Based on the event's nature, one event can be translated into 1 up to 7 tokens. Each token derived from the same event shares its timestamp. The final step involves representing time gaps between events by inserting time-interval tokens. If the time difference between events is less than 5 minutes—the minimum value represented by the token for the shortest time interval—no token is added. After adding interval-tokens, timestamps are stripped from the timeline.
    }
    \label{fig:token-schema}
\end{figure}

\subsection{Patient health timelines (PHTs), tokenization}

\begin{figure}[ht]
    \centering
    \includegraphics[width=1\linewidth]{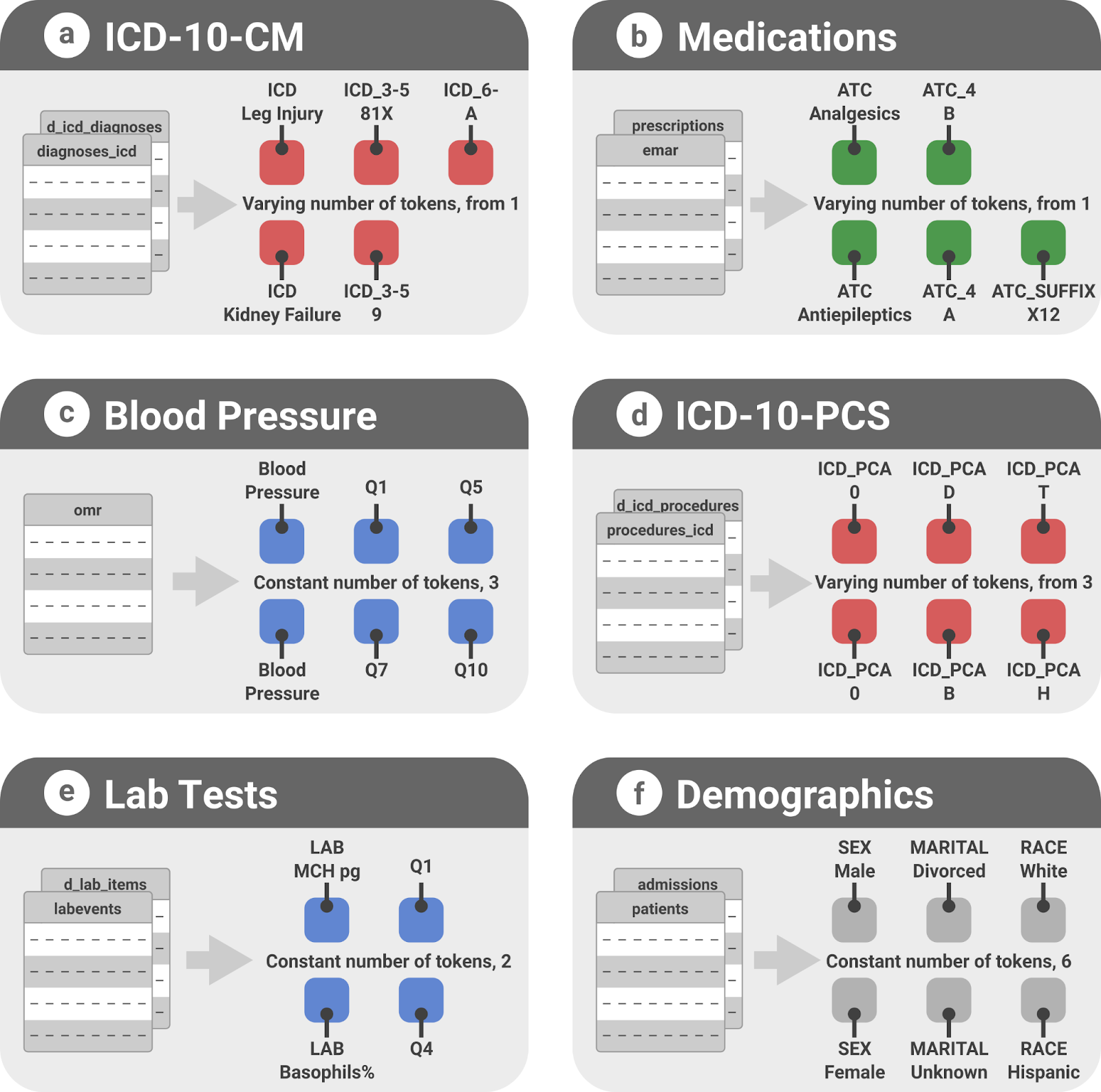}
    \caption{
        \textbf{Various examples of information encoding via tokens.}
        Depending on the ICD-10-CM code, 1 to 3 tokens are utilized for representation, with the first token corresponding to the code's first three characters, the fourth and fifth characters possibly represented by another token, and an optional third token for the remaining characters in the ICD code. (b) Medications, coded by ATC codes, are similarly encoded by 1 to 3 tokens based on the specificity of the code, with the first token representing the first three characters, the second for the next two, and the third for the remaining characters. (c) Blood pressure measurements are consistently encoded using three tokens: one to indicate the BP measurement and two quantile tokens for systolic and diastolic pressure values, respectively. (d) ICD-PCS codes may be represented by up to seven tokens, with each token denoting one character of the code. (e) Lab tests are depicted by a token that describes the type of test followed by a quantile token for the test's numerical value. Finally, (f) demographics are depicted which are part of static tokens, always positioned at the beginning of the PHT.
    }
    \label{fig:sup-5}
\end{figure}

The core concept behind ETHOS is the Patient Health Timeline (PHT), as depicted in Figure~\ref{fig:pipeline}. The fundamental component of the PHT is the token, which represents a distinct unit of information occurring within the patient's health timeline. To construct the PHT, we gathered all pertinent data from tables 1 to 12 of the MIMIC-IV database, as detailed in the Data section. We arranged this data chronologically based on timestamps, as shown in Figure~\ref{fig:token-schema}, into a chronological sequence of health-related events for each patient. These events were timestamped with a floating-point number in 64-bit precision to denote the patient's age at the time of occurrence of the event. Subsequently, events from the MIMIC-IV tables were converted into tokens. Each event was represented by 1 to 7 tokens to encapsulate information about the event, as illustrated in Figure~\ref{fig:sup-5}a. We crafted this encoding process to ensure each token conveys specific, meaningful information, with examples in Figure~\ref{fig:sup-5}c-f. A comprehensive list of token encodings within the PHT is available in the supplementary material. The final step of tokenization involved the insertion of time-interval tokens to represent the intervals between events, depicted in Figure~\ref{fig:token-summary}c. We employed 13 different time-interval tokens to represent the intervals. No interval token was inserted if the duration between tokens was less than 5 minutes. Typically, a single time-interval token was placed between other types of tokens unless the interval exceeded one year. In such cases, multiple 6-month tokens were used to approximate the actual interval. For example, an interval of 1.4 years was represented by three 6-month tokens, while four 6-month tokens represented 1.76 years. One interval-tokens were inserted the exact time of events was dropped from PHTs.

The patient's age and the commencement date of the PHT were represented using the same token set. We used 20 distinct tokens to denote age intervals such as 0-5 years, 5-10 years, and so forth. For instance, to encode information about a 46-year-old patient with PHT beginning in 1982, we inserted a "45\_50 years" token at the 4th position in the PHT. To signify the year 1982, we used a "15\_20 years" token at the 5th position of the PHT, considering 1970 as the baseline year. We emphasize that age and the commencement of the PHT are encoded in five-year intervals, given that health status typically does not undergo rapid changes with age, making finer granularity unnecessary. However, we plan to scrutinize these assumptions in subsequent research. The token denoting the commencement of the PHT delineates the temporal context of the medical data—identifying whether it corresponds to earlier medical practices (e.g., 1990s), contemporary practices, or periods in between. Using tokens with a precision of five years is done under the premise that technological and methodological progress within the medical field does not advance at a pace that justifies the necessity for time intervals more granular than five-year spans. Pertinent to the MIMIC dataset, the obfuscation of actual dates through uniform random adjustments for each patient—a measure implemented to safeguard privacy—compromises the utility of this temporal information for ETHOS, as it obscures the precise date of the start of PHT. However, the absence of precise reference dates is less critical, given that the entire dataset was collected over a relatively brief period, from 2008 to 2019 (\cite{johnson_mimic-iv_2023}).

As mentioned previously, token locations within the timeline are contingent upon the temporal occurrence of events. Nonetheless, certain data elements are temporally invariant, or at least presented as such within the MIMIC-IV database. In our implementation, we designate six static tokens to encapsulate the information encoded in these static data elements. Although, in reality, some of these variables may change over time, they are represented as invariable constants in the MIMIC database. We encoded this information in the six static tokens exactly as recorded in the MIMIC dataset. These include gender, marital status, race, body mass index (BMI), birth date, and the start date of the timeline. While PHTs have the potential to extend to hundreds of thousands of tokens, our current methodology utilizes a maximum of 2048 subsequent tokens within the transformer model context, as elaborated in the "Methods: ETHOS Training" section. To accommodate invariant data, we substitute the initial six tokens of the 2048-token context with static information tokens, where the sixth token demarcates the temporal juncture of the seventh token, which is the first token of the actual timeline. Although the transformer architecture inherently facilitates the inclusion of static data via its encoder component and cross attention module3, we opted for a more streamlined approach as described, deferring the integration of an encoder implementation to future endeavors where more substantial time-invariant data like genetics is used.

\begin{figure}[ht]
    \centering
    \begin{subfigure}[b]{0.47\linewidth}
        \centering
        \includegraphics[width=\textwidth]{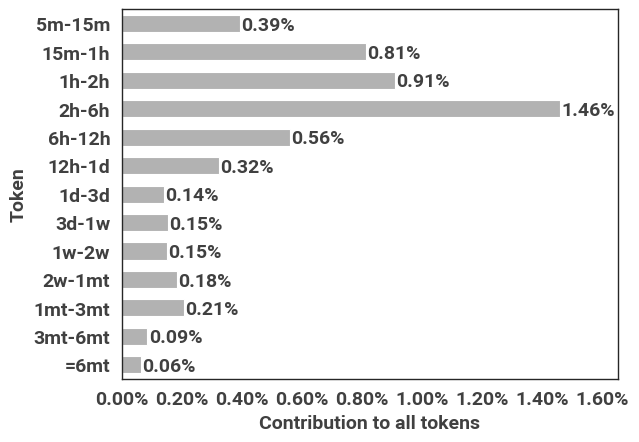}
    \end{subfigure}
    \hfill
    \begin{subfigure}[b]{0.47\linewidth}
        \centering
        \includegraphics[width=\textwidth]{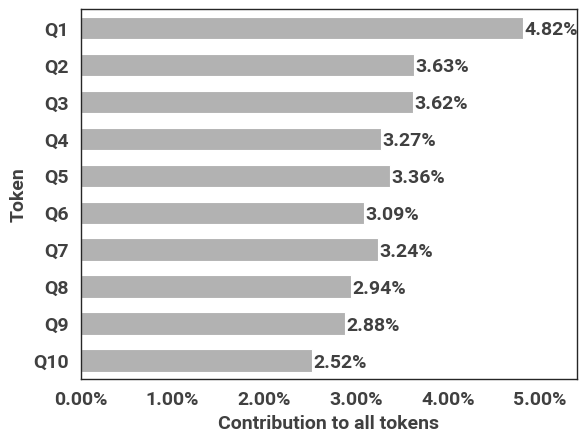}
    \end{subfigure}
    \caption{\textbf{The variability in the representation of time intervals and quantile tokens within the tokenized dataset.} The longest used interval was 6 months as it was relatively rare to have intervals much longer than that. Notably, for tests with discrete ordered outputs, a reduced number of quantiles was employed, resulting in a non-uniform frequency distribution across these specific categories.}
    \label{supfig:4}
\end{figure}

Medical encounters yield a plethora of numerical data. We employ a quantile-based tokenization strategy to process continuous numerical values, such as blood pressure readings or cholesterol levels. Specifically, all numerical values are transformed into integers representing the quantile to which each value corresponds. Quantile ranges were determined using the training dataset, where histograms of all numerical values were generated and subsequently divided into quantiles. We chose to utilize ten quantiles, a decision aimed at striking a balance between the need for precise representation of numerical data and the clinical reality that significant changes in health indicators often manifest as relatively large variations, such as shifts of 10 or 20 percent (Figure~\ref{supfig:4}). This rationale underpins our selection of ten quantiles for tokenization.

In our study, Diagnosis-Related Group (DRG) codes for each inpatient stay were utilized, despite the absence of assigned times when they were created in the MIMIC tables. Given that a DRG code is assigned after or during discharge, we positioned it after a trio of tokens representing discharge-related information: the discharge token, a quantile token indicating the length of the hospital stay, and a token specifying the discharge destination (e.g., home). Additionally, we incorporated data from MIMIC regarding the initial SOFA score for ICU patients, placing this token after the patient's admission-to-the-ICU token, along with a token denoting the ICU type. Given that the SOFA score in the dataset ranges from 0 to 23 (with the score of 24 never appearing), we uniformly map scores from 0-23 across 1-10 quantiles. Consequently, in quantile Q1, SOFA scores of 0, 1, and 2 (average of 1) are included, while quantile Q2 encompasses SOFA scores of 3 and 4 (average of 3.5), and this pattern continues accordingly.

ETHOS operates as a causal network. It relies solely on information available up to the time being considered in making predictions. Consequently, to ensure causality, actual values of DRG codes and SOFA scores are not employed during inference; instead, predictions of these values are used. This principle ensures that future-obtained information does not influence the prediction of yet-to-occur events. In essence, if tokens are integrated into the timeline based on their approximate occurrence time, their actual values must not be utilized for inference purposes, or they are placed in the timeline far in the future to ensure they are inserted after they occurred.

\begin{figure}[ht]
    \centering
    \includegraphics[width=1\linewidth]{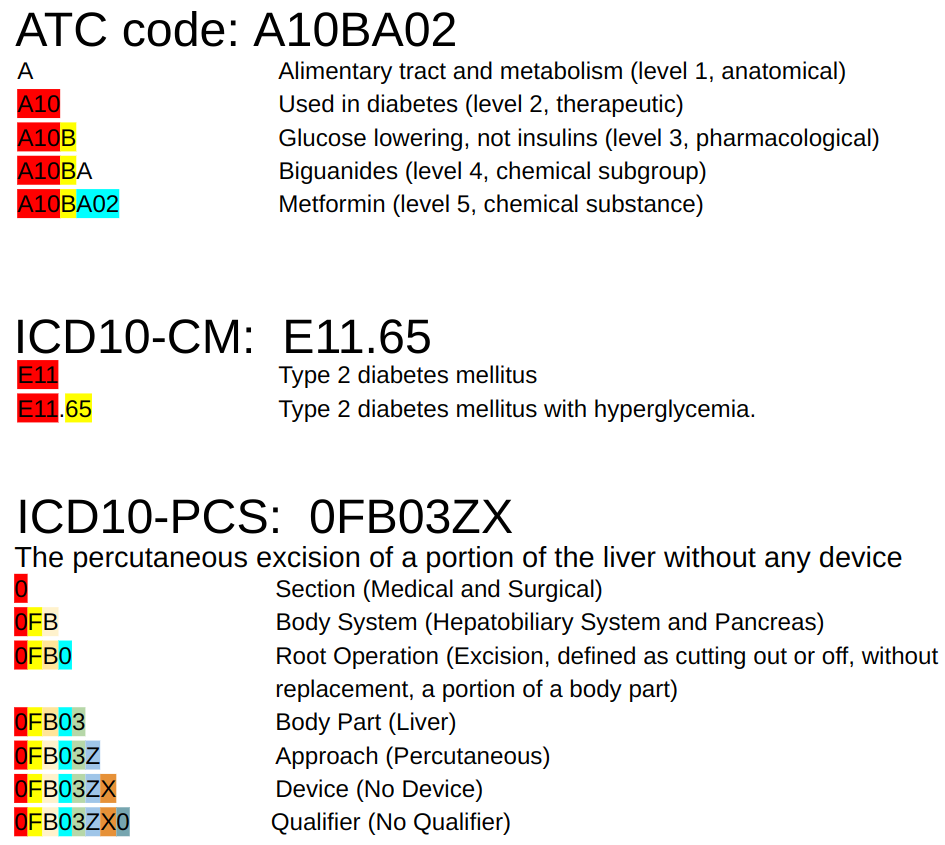}
    \caption{
        \textbf{Tokenization schema of ATC, ICD10-CM, and ICD10-PCS.}
        Different colors correspond to parts of the codes encoded by different tokens in examples below.
    }
    \label{supfig:1}
\end{figure}

For the tokenization of drugs, whether administered or prescribed, we utilized the ATC classification system due to its hierarchical, tree-like structure (Figure~\ref{supfig:1}). Each ATC code, comprising up to seven characters, was encoded using up to three sequential tokens: the first token for the initial three characters, the second for the subsequent character, and the third optional token, for the remaining suffix. Similarly, ICD-10-CM codes were encoded with three tokens: the first representing the first three characters of the code, the next two by the second token, and the final token capturing the code's remaining suffix. For ICD-10-PCS codes, each character in the seven-character code was represented by a distinct token. The rationale behind such tokenization is that the initial characters in those coding schemes denote specific classes of drugs and diseases or procedures, which are interpretable and have distinct meanings which we anticipated to be important for the network's self-attention mechanisms. Looking ahead, our approach, which assigns well-defined meanings to each token, will be crucial for refining attention mechanisms and enhancing the model's explainability. This method ensures that individual tokens contribute significantly to the interpretability of the network's outcomes. For more information on the tokenization process applied to MIMIC data in our analysis, as well as examples of Patient Health Timelines (PHTs), readers are directed to Table~\ref{ltab:t1} and Table~\ref{ltab:t2} where we present real PHTs used in this work with annotations. A summary of all tokenized components of the MIMIC dataset is in Table~\ref{tab:events}.

\subsection{ETHOS training}

\begin{figure}[ht]
    \centering
    \includegraphics[width=1\linewidth]{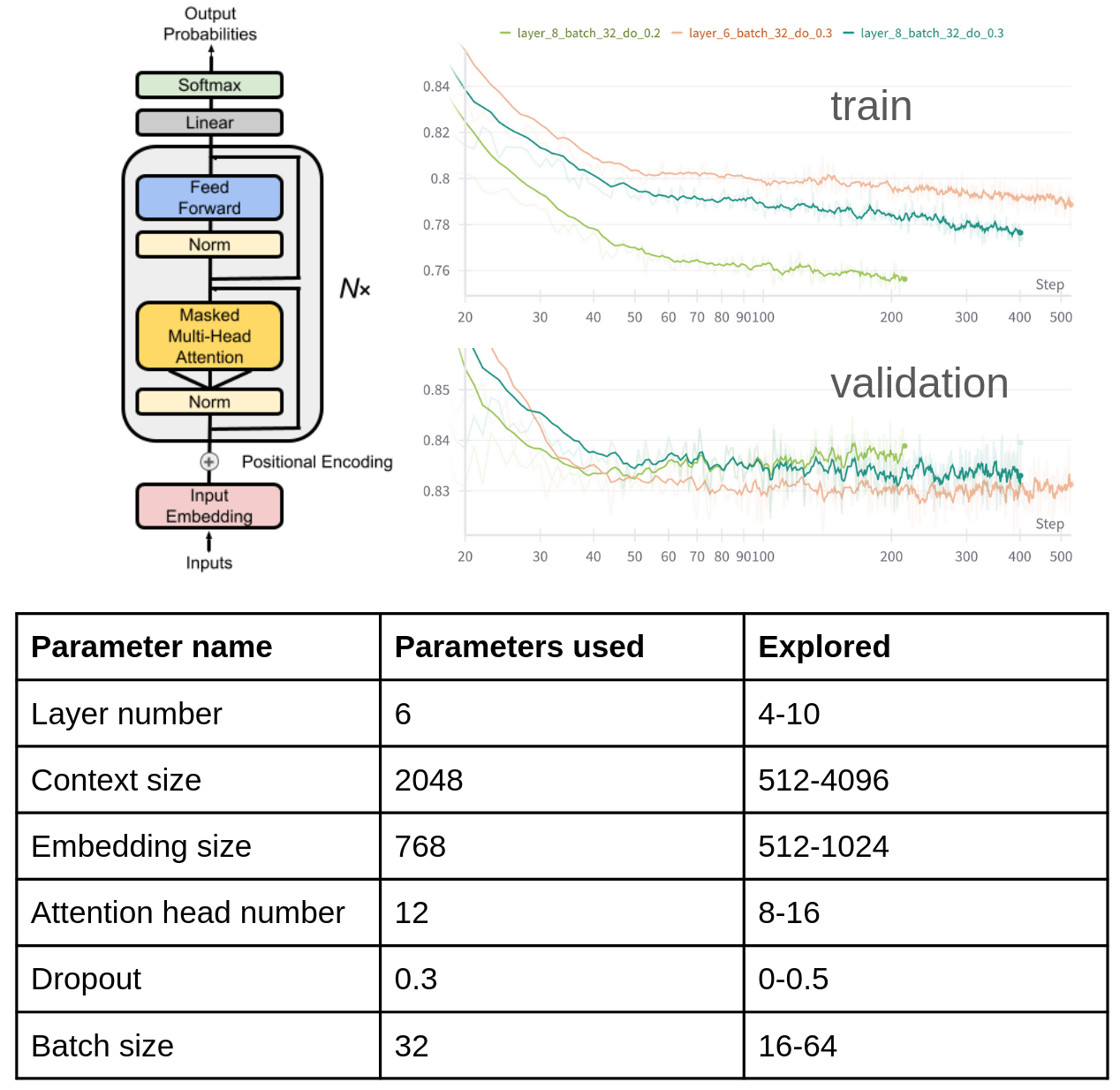}
    \caption{
        \textbf{Architecture of the Transformer Decoder Model employed in ETHOS.}
        We closely mirrored the original decoder design by~\cite{vaswani_attention_2017}. This iteration of the model was initially sized in alignment with GPT-2 (\cite{brown_language_2020}), reflecting a comparable scale of training data. Included are select training traces and a detailed account of the operational model, complete with parameter specifications outlined in the accompanying table.
    }
    \label{supfig:2}
\end{figure}

We employ a model inspired by the decoder architecture of the transformer~\cite{vaswani_attention_2017}, drawing parallels between tokenized text in Natural Language Processing (NLP) and our approach to tokenizing PHTs. We based our model development on Andrej Kapathy’s implementation of GPT-2~\footnote{\url{https://github.com/karpathy/nanoGPT}}. The design choice slightly varies from the original transformer paper, because instead of using fixed sinusoidal positional encodings, it utilizes learneable position embeddings that are added to the token embeddings at the stage where tokens are converted to their corresponding embeddings. The ETHOS model's training begins by synthesizing a dataset from existing patient records. Each patient's PHT is ended with a "End of timeline" token, and then they are concatenated, creating a single long sequence of tokens for the training. Similarly to generative LLM, ETHOS is trained to predict a single token based on the context of preceding ones. Given the large data scale and model complexity, this phase is resource-intensive similar to methods for training used for NLP transformers used in LLMs (\cite{vaswani_attention_2017, thirunavukarasu_large_2023}). We estimated that the size of the network training task that we face with ETHOS is similar to GPT-2~\cite{brown_language_2020}, and therefore we used the size of the transformer used in that network as a starting point (details on the hyperparameter search and choice can be seen in Figure~\ref{supfig:2}). We made heuristic adjustments to the size of the network to optimize the value of the loss function. Further details on our training methodology of transformers are provided in~\cite{brown_language_2020} and for our implementation in supplementary material and full complete code published on GitHub~\footnote{\url{https://github.com/ipolharvard/ethos-paper}}.

\subsection{Evaluation of Clinical Outcomes and Tasks Using ETHOS}

The experiments were chosen so the results can be compared to the work of others in terms of the estimation of inpatient mortality and readmission on MIMIC data. Patients in the MIMIC were randomly divided into training and testing groups, with splits of 90\%/10\% (Table~\ref{tab:demographics}).

The chance of inpatient mortality was assessed at the time of admission for all inpatient stays for patients in the test set unless the discharge day was unknown. This was performed by the generative process that began with the admission token and ended upon generating a discharge or death token, repeating this cycle 20 times. The 'N', representing the number of times a death token was generated first, was divided by 20 to estimate the chance of inpatient mortality. Similarly, the likelihood of ICU mortality was computed for the MIMIC dataset, with an additional experiment conducted where predictions were made starting 24 hours after ICU admission, rather than at the point of ICU admission. In the same simulation, the LOS in the ICU was estimated by aggregating the time-interval tokens generated in the simulated timeline until the discharge token appeared. Instances where the patient died in the ICU during the simulation were excluded from the LOS calculation. We opted for 20 repetitions, yielding 21 unique probability estimators, which were adequate for constructing robust Receiver Operating Characteristic (ROC) curves yielding excellent Gaussian fits (Figure~\ref{fig:results}). Nevertheless, alternative repetition counts may also be employed.

To calculate the probability of 30-day inpatient readmission, the generation of fPHTs commenced at the discharge token from inpatient stays and ceased upon the appearance of either a new admission or death token or when the cumulative time tokens generated exceeded 30 days. The simulation was repeated 20 times. The probability of 30-day readmission was then derived as M/20, where 'M' is the count of terminations occurring because of patient new admission tokens across the 20 repetitions.

In our approach, tasks are accomplished by simulating future patient health timelines. Yet, ETHOS offers additional methods for deriving insights, two of which we illustrate here. For instance, in the construction of PHTs following each ICU admission, a sequence is created starting with a token that identifies the type of ICU, followed by a SOFA score token, and then by a Q token that signifies the actual SOFA score on the first day. We predict the SOFA score using SOFA Q node probabilities as generated by ETHOS and the mean SOFA score per quantile as assigned during tokenization (Figure~\ref{fig:results-sofa-drg}a).

The exact timing of the 1-day SOFA score assessment is not specified in the dataset, leading to a potential causality issue by inserting the SOFA score immediately after admission, as it relies on data acquired subsequently. During the model's training phase, ETHOS permits this apparent causality violation. However, such true values of 1-day SOFA scores, not available at the moment of ICU admission, are not used for simulating future timelines during inference to prevent causality violation during inference. Instead, these scores are predicted from prior information, as demonstrated in our study. This feature of ETHOS enables the inclusion of information with indeterminate timing.

Another distinctive inference capability facilitated by ETHOS is DRG class estimation. As illustrated in Figure~\ref{fig:results-sofa-drg}c, the token denoting the DRG class is consistently positioned following the discharge token and a Q token specifying the length of hospital stay. With 771 unique tokens available for this purpose, we infer the actual class by generating a probability array in the final network layer of the transformer for the DRG token. This array is then utilized to predict the classification's top-1 and top-2 accuracy metrics.

\subsection{Statistical Analysis}

The performance of classification algorithms of binary tasks was assessed using Receiver Operating Curve Analysis (ROC). The ROC curves were fitted to experimental points using Gaussian models with unequal variances for binary hypotheses (code provided). Values of Areas Under Curves (AUCs) and 95\% confidence intervals (CI) were calculated using bootstrapping (code provided). For multiclass classification (DRG task), we used top-1 and top-2 accuracy. We used mean absolute error (MEA) for the regression tasks to indicate prediction fidelity with 95\% confidence intervals estimated using bootstrapping. Python numpy and scikit-learn were used.

\subsection{Comparison of ETHOS to existing methods}

\begin{figure}[ht]
    \centering
    \includegraphics[width=1\linewidth]{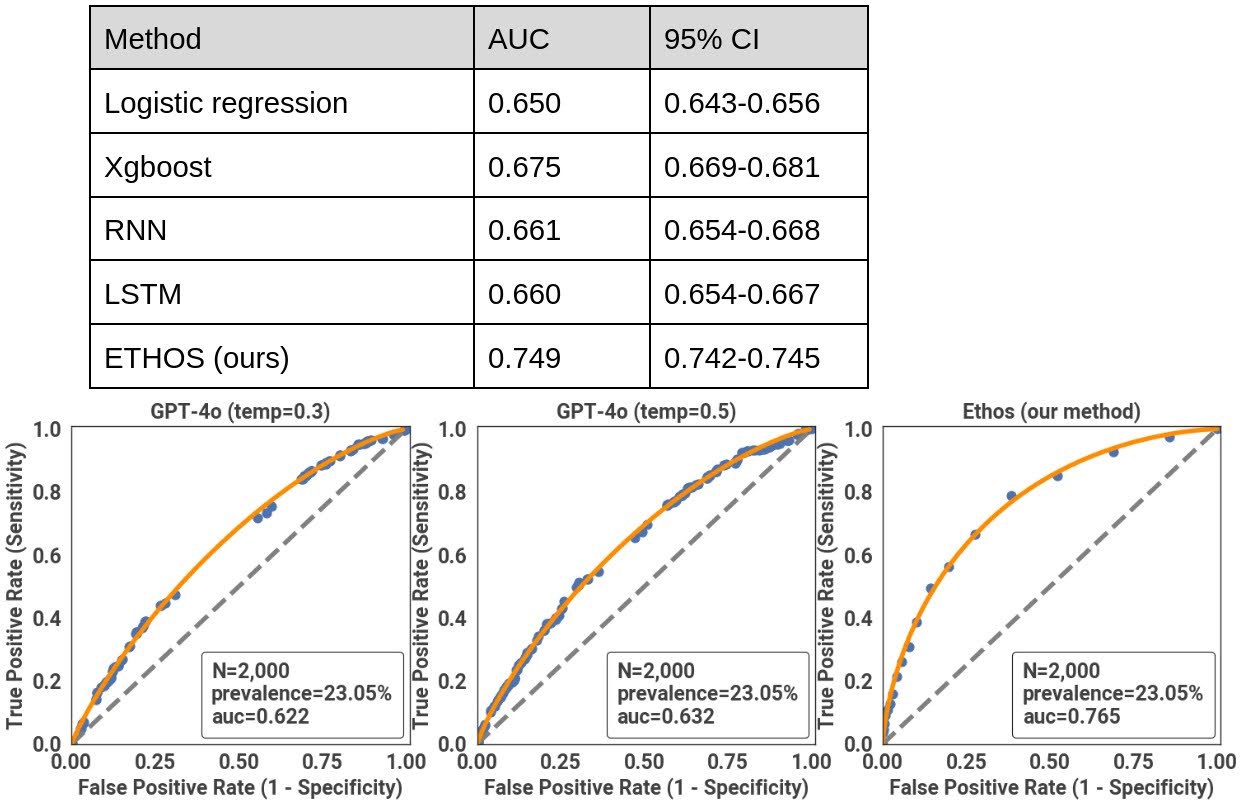}
    \caption{
        \textbf{Comparison of Ethos to existing methods.}
        Features were derived from patient hospitalization data following \cite{tang_predicting_2023}. Applying their algorithm was impractical due to the need to compute similarity scores for each admission pair in our larger dataset (400,000 vs. 14,500 admissions), which exceeded our 2TB RAM capacity. We therefore used only their data preprocessing and feature extraction methods. Modified code from Tang et al.’s repository is available on GitHub (\url{www.github.com/ipolharvard/readmit-stgnn}). For non-temporal models like Logistic Regression and XGBoost, time-varying features were summarized using statistical measures, and the day of admission was included as a feature. ETHOS was compared to GPT-4o at temperatures 0.3 and 0.5. We created a prompt directing the model to analyze 2048 tokens and calculate readmission probability for 2000 cases. ETHOS outperformed both GPT-4o variants.
    }
    \label{fig:sup-6}
\end{figure}

Employing the data segmentation as detailed in Table~\ref{tab:demographics}, we evaluated traditional algorithms for predicting 30-day hospital readmission rates and juxtaposed these outcomes with those obtained via ETHOS. The features used in Figure~\ref{fig:sup-6} were culled from data accrued during the patient's hospitalization, adhering to the feature derivation methodology outlined by~\cite{tang_predicting_2023}. Attempting to apply the algorithm devised by the authors to our dataset presented challenges, notably due to the Graph Neural Network (GNN) implementation by Tang et al., which necessitates the computation of a similarity score for each pair of admissions. Given the significantly larger volume of admissions in our dataset—approximately 400,000, in stark contrast to the 14,500 reported by Tang et al.—this task proved impractical on a compute node with 2TB of RAM, defying all efforts to achieve it within a reasonable timeframe. Consequently, we limited our application to the data preprocessing and feature extraction segments of Tang et al.’s methodology. The adapted and modified code from Tang et al.’s repository, which we cloned for feature extraction, is accessible on GitHub~\footnote{\url{https://github.com/ipolharvard/readmit-stgnn}}. For models unsupportive of temporal sequence analysis, such as Logistic Regression and XGBoost, we modified the approach to handle time-varying features by consolidating them over time. This entailed distilling the minimum, first quartile, median, third quartile, and maximum values of dynamically changing features. Furthermore, we integrated the day of admission as a unique feature to retain an element of temporal dimension within the dataset. In Figure~\ref{fig:sup-6} ETHOS was compared to one of the leading proprietary LLM models - GPT-4o in two temperature variants: 0.3 and 0.5. We constructed a comprehensive prompt that directs the model to analyze a timeline of 2048 tokens and calculate the probability of patient readmission for 2000 cases from the test set. This prompt is structured into four distinct parts: task instructions, a basic patient description corresponding to ETHOS's static information, PHT and a detailed description of subgroups of tokens and their identification. The complete codebase for this experiment including the prompt design is accessible on GitHub~\footnote{\url{https://github.com/ipolharvard/ethos-paper/blob/master/notebooks/llm_readmission_task.ipynb}}. ETHOS significantly outperforms both variants of GPT-4o for the same subset of testing samples.

\section*{Additional information}

\paragraph{Data Availability} The MIMIC-IV dataset is publicly available at \url{https://physionet.org/content/mimiciv/2.2}.

\paragraph{Code Availability} The code, ETHOS model weights used for all inferences, results of inferences, scripts to generate numerical results for all aspects of this study for the MIMIC-IV dataset are made publicly available at \url{https://github.com/ipolharvard/ethos-paper}. In our experiments, we used Python 3.10, and the following open-source libraries: torch=2.3.0, joblib=1.4.2, tqdm=4.66.4, colorlog=6.8.2, h5py=3.11.0, pandas=2.2.2, numpy=1.26.4, pyarrow=16.1.0, click=8.1.7.

\paragraph{Author Contribution} AS and PR conceptualized the work. AS, PR, YJ, AES designed the study. PR, AS performed the coding and the experiments. YJ, AS conducted the literature search. DB, AES, QL, JW provided advisory support for the project. PR and AS prepared the initial draft of the manuscript, with all authors actively participating in the refinement and finalization of the manuscript through comprehensive review and contributions. AS supervised the project.

\paragraph{Competing Interests} YJ is currently also affiliated with Verily life science, SSF, CA. The other authors declare no competing interests.

\bibliographystyle{unsrtnat}
\bibliography{references}

\section*{Supplementary Information}

\SupplementaryMaterials

\begin{table}[ht]
    \centering
    \begin{tabular}{lrrr}
        \toprule
        Characteristics & Train/Validation & Test & Total \\
        \midrule
        Patient number & 241,015 & 26,758 & 267,773 \\
        \midrule
        \textbf{Age, years, mean (std)} & 50.27 (20.76) & 50.15 (20.80) & 50.25 (20.77) \\
        \midrule
        \textbf{Gender} & & & \\
        \hspace{1em} Female & 130,115 & 14,473 & 144,588 \\
        \hspace{1em} Male & 110,900 & 12,285 & 123,185 \\
        \midrule
        \textbf{Race} & & & \\
        \hspace{1em} White & 109,274 & 12,090 & 121,364 \\
        \hspace{1em} Unknown & 87,420 & 9,724 & 97,144 \\
        \hspace{1em} Black & 21,048 & 2,361 & 23,409 \\
        \hspace{1em} Hispanic & 8,991 & 1,014 & 10,005 \\
        \hspace{1em} Other & 7,506 & 823 & 8,329 \\
        \hspace{1em} Asian & 6,776 & 746 & 7,522 \\
        \midrule
        \textbf{Marital Status} & & & \\
        \hspace{1em} Unknown & 84,533 & 9,369 & 93,902 \\
        \hspace{1em} Married & 70,297 & 7,649 & 77,946 \\
        \hspace{1em} Single & 60,822 & 6,910 & 67,732 \\
        \hspace{1em} Widowed & 15,068 & 1,719 & 16,787 \\
        \hspace{1em} Divorced & 10,295 & 1,111 & 11,406 \\
        \bottomrule
    \end{tabular}
    \caption{\textbf{Demographic characteristics of the dataset.} The characteristics are reported at the time of the first hospital admission and used by ETHOS.}
    \label{tab:demographics}
\end{table}

\begin{table}[ht]
    \centering
    \begin{tabular}{llll}
        \toprule
        Event or Information & Token \# & \# of Unique Tokens & Notes \\
        \midrule
        Sex & 1 & 2 & Always at position 1 in PHT \\
        Race & 1 & 6 & Always at position 2 in PHT \\
        Marital status & 1 & 5 & Always at position 3 in PHT \\
        BMI & 1 & 1 (+10) & Always at position 4 in PHT \\
        Age at the start of timeline & 1 & 21 & Always at position 5 in PHT \\
        Years from 1970 at the start of timeline & 1 & 21 & Always at position 6 in PHT \\
        Deaths & 1 & 1 & N/A \\
        Emergency Department Stays & 2 & 2 (+10) & Admission/discharge + Q \\
        Inpatient Stays & 6 & 18 (+10) & N/A \\
        Transfers & 1 & 19 & 19 different transfer types \\
        ICU Stays & 4 & 11 (+10) & N/A \\
        Diagnoses - ICD-10-CM & 1-3 & 2936 & N/A \\
        Procedures - ICD-10-PCS & 1-7 & 35 & N/A \\
        Blood Pressure & 3 & 1 (+10) & BP token + 2Qs (sys+dia) \\
        Administered Medications & 1-3 & 312 & By tokenizing ATC code \\
        Lab tests & 2 & 200 (+10) & Used most frequent 200 lab tests + Q \\
        SOFA & 2 & 1 (+10) & Always follow ICU admin \\
        DRG & 1 & 772 & \makecell[l]{Always follow Q token after inpatient \\ discharge. 771 DRG classes + unknown} \\
        \bottomrule
    \end{tabular}
    \caption{\textbf{Medical events extracted from the MIMIC database and an overview of the token encodings utilized.}}
    \label{tab:events}
\end{table}

\clearpage
\begin{longtable}{ll}
    \caption{Example of a tokenized timeline with a hospital admission}\label{ltab:t1}\\
    \toprule
    \textbf{Patient Health Timeline} & \textbf{Notes} \\
    \midrule
    \endfirsthead

    \multicolumn{2}{c}%
    {\tablename\ \thetable\ -- \textit{Continued from previous page}} \\
    \midrule
    \textbf{Patient Health Timeline} & \textbf{Notes} \\
    \midrule
    \endhead

    \midrule 
    \multicolumn{2}{r}{\textit{Continued on next page}} \\
    \endfoot

    \bottomrule
    \endlastfoot
    
    ED\_ADMISSION\_START & Admission to Emergency Department \\
    \_2h-6h & Time interval between 2h and 6h \\
    LAB\_pH\_units & Lab test of pH and the unit is units \\
    \_Q4 & Quantile token referring to the lab test of pH \\
    LAB\_Protein\_mg/dL & \\
    \_Q4 & \\
    LAB\_RBC\_\#/hpf & \\
    \_Q1 & \\
    LAB\_Specific Gravity & \\
    \_Q2 & \\
    LAB\_WBC\_\#/hpf & \\
    \_Q1 & \\
    LAB\_Alanine Aminotransferase (ALT)\_IU/L & \\
    \_Q3 & \\
    LAB\_Albumin\_g/dL & \\
    \_Q3 & \\
    LAB\_Alkaline Phosphatase\_IU/L & \\
    \_Q1 & \\
    LAB\_Anion Gap\_mEq/L & \\
    \_Q5 & \\
    LAB\_Asparate Aminotransferase (AST)\_IU/L & \\
    \_Q8 & \\
    LAB\_MCHC\_g/dL & \\
    \_Q9 & \\
    LAB\_MCV\_fL & \\
    \_Q7 & \\
    LAB\_Monocytes\_\% & \\
    \_Q7 & \\
    LAB\_Neutrophils\_\% & \\
    \_Q3 & \\
    LAB\_Platelet Count\_K/uL & \\
    \_Q3 & \\
    LAB\_RDW\_\% & \\
    \_Q1 & \\
    LAB\_Red Blood Cells\_m/uL & \\
    \_Q8 & \\
    LAB\_White Blood Cells\_K/uL & \\
    \_Q5 & \\
    LAB\_Absolute Basophil Count\_K/uL & \\
    \_Q5 & \\
    LAB\_Absolute Eosinophil Count\_K/uL & \\
    \_Q4 & \\
    LAB\_Absolute Monocyte Count\_K/uL & \\
    \_Q6 & \\
    \_6h-12h & \\
    INPATIENT\_ADMISSION\_START & Token indicating an admission to the hospital \\
    TYPE\_OBSERVATION & The type of the admission is observation \\
    INSURANCE\_MEDICARE & The patient’s insurance is MEDICARE \\
    \makecell[l]{ICD\_Other symptoms and signs involving cognitive \\ functions and awareness} & \makecell[l]{The primary diagnosis at the beginning of the hospital stay\\ is Altered mental status, unspecified (R4182), which is \\ broken down into two tokens: R41 and 82} \\
    ICD\_4-5\_82 & \\
    ICD\_Dorsalgia & \\
    ICD\_4-5\_16 & \\
    ICD\_Malaise and fatigue & \\
    ICD\_4-5\_1 & \\
    ICD\_Personal history of certain other diseases & \\
    ICD\_4-5\_73 & \\
    ICD\_Other hypothyroidism & \\
    ICD\_4-5\_9 & \\
    ICD\_Type 2 diabetes mellitus & \\
    ICD\_4-5\_21 & \\
    ICD\_Dorsalgia & \\
    ICD\_4-5\_5 & \\
    ICD\_Abnormalities of gait and mobility & \\
    ICD\_4-5\_2 & \\
    TRANSFER\_MED & Transfer to a different care unit - MED \\
    \_6h-12h & \\
    ATC\_stomatological preparations & \\
    ATC\_4\_A & \\
    ATC\_SUFFIX\_D05 & \\
    ATC\_diuretics ATC\_4\_A & \\
    ATC\_4\_A & \\
    ATC\_SUFFIX\_A03 & \\
    ATC\_agents acting on the renin-angiotensin system & \\
    ATC\_4\_A & \\
    ATC\_SUFFIX\_A03 & \\
    \_6h-12h & \\
    ED\_ADMISSION\_END & Discharge from Emergency Department \\
    \_Q10 & \makecell[l]{Quantile referring to the length of the stay in Emergency \\ Department based on all stays in the data} \\
    INPATIENT\_ADMISSION\_END & Discharge from the hospital \\
    \_Q2 & Quantile referring to the length of the stay in the hospital \\ based on all stays in the data \\
    DISCHARGED\_UNKNOWN & The reason of discharge \\
    UNKNOWN\_DRG & DRG assigned to the hospital stay \\
    \_=6mt & Time interval of 6 months \\
    \bottomrule
\end{longtable}

\begin{longtable}{ll}
    \caption{Example of a tokenized timeline with an admission to ICU}\label{ltab:t2}\\
    \toprule
    \textbf{Patient Health Timeline} & \textbf{Notes} \\
    \midrule
    \endfirsthead

    \multicolumn{2}{c}%
    {\tablename\ \thetable\ -- \textit{Continued from previous page}} \\
    \midrule
    \textbf{Patient Health Timeline} & \textbf{Notes} \\
    \midrule
    \endhead

    \midrule 
    \multicolumn{2}{r}{\textit{Continued on next page}} \\
    \endfoot

    \bottomrule
    \endlastfoot
    
    ICU\_STAY\_START & Admission to ICU token \\
    Surgical Intensive Care Unit (SICU) & Type of ICU \\
    SOFA & Indicates SOFA quantile follows \\
    \_Q1 & Q indicating quantile of SOFA score \\
    \_15m-1h & time-interval \\
    LAB\_Sodium\_mEq/L & Lab test results \\
    \_Q7 & \\
    LAB\_Urea Nitrogen\_mg/dL & \\
    \_Q3 & \\
    LAB\_L\_no\_unit & \\
    \_Q4 & \\
    \_15m-1h & time-interval \\
    LAB\_Base Excess\_mEq/L & \\
    \_Q6 & \\
    LAB\_Calculated Total CO2\_mEq/L & \\
    \_Q8 & \\
    LAB\_pCO2\_mm Hg & \\
    \_Q7 & \\
    LAB\_pH\_units & \\
    \_Q7 & \\
    LAB\_pO2\_mm Hg & \\
    \_Q5 & \\
    \_2h-6h & \\
    ATC\_anti-asthmatics & Medications start and administered with specified time intervals \\
    ATC\_4\_A & \\
    ATC\_SUFFIX\_C02 & \\
    \_5m-15m & \\
    ATC\_analgesics & \\
    ATC\_4\_B & \\
    ATC\_SUFFIX\_E01 & \\
    \_2h-6h & \\
    ATC\_drugs for acid related disorders & \\
    ATC\_4\_B & \\
    ATC\_SUFFIX\_C02 & \\
    \_1h-2h & \\
    ATC\_antithrombotic agents & \\
    ATC\_4\_A & \\
    ATC\_SUFFIX\_B01 & \\
    \_15m-1h & \\
    ATC\_beta blocking agents & \\
    ATC\_4\_A & \\
    ATC\_SUFFIX\_B02 & \\
    \_1h-2h & \\
    ATC\_mineral supplements & Patient is medicated until the end of ICU stay \\
    ATC\_4\_C & \\
    ATC\_SUFFIX\_C02 & \\
    \_15m-1h & \\
    ATC\_laxatives & \\
    ATC\_4\_& \\
    A ATC\_SUFFIX\_D04 & \\
    \_1h-2h & \\
    ATC\_analgesics & \\
    ATC\_4\_B & \\
    ATC\_SUFFIX\_E01 & \\
    \_15m-1h & \\
    ATC\_antipruritics, anesthetics, etc. & \\
    ATC\_4\_& \\
    A ATC\_SUFFIX\_A32 & \\
    LAB\_Hematocrit\_\% & Here lab test results are reported \\
    \_Q7 & \\
    LAB\_Hemoglobin\_g/dL & \\
    \_Q6 & \\
    LAB\_MCH\_pg & \\
    \_Q8 & \\
    LAB\_MCHC\_g/dL & \\
    \_Q5 & \\
    LAB\_MCV\_fL & \\
    \_Q9 & \\
    LAB\_Platelet Count\_K/uL & \\
    \_Q3 & \\
    LAB\_RDW\_\% & \\
    \_Q4 & \\
    LAB\_Red Blood Cells\_m/uL & \\
    \_Q5 & \\
    LAB\_White Blood Cells\_K/uL & \\
    \_Q4 & \\
    LAB\_RDW-SD\_fL & \\
    \_Q6 & \\
    LAB\_INR(PT)\_no\_unit & \\
    \_Q4 & \\
    LAB\_PT\_sec & \\
    \_Q5 & \\
    LAB\_PTT\_sec & \\
    \_Q2 & \\
    LAB\_Anion Gap\_mEq/L & \\
    \_Q4 & \\
    LAB\_Bicarbonate\_mEq/L & \\
    \_Q4 & \\
    LAB\_Calcium, Total\_mg/dL & \\
    \_Q3 & \\
    LAB\_Chloride\_mEq/L & \\
    \_Q5 & \\
    LAB\_Creatinine\_mg/dL & \\
    \_Q2 & \\
    LAB\_Glucose\_mg/dL & \\
    \_Q8 & \\
    LAB\_H\_no\_unit & \\
    \_Q9 & \\
    LAB\_I\_no\_unit & \\
    \_Q2 & \\
    LAB\_Magnesium\_mg/dL & \\
    \_Q2 & \\
    LAB\_Phosphate\_mg/dL & \\
    \_Q1 & \\
    LAB\_Potassium\_mEq/L & \\
    \_Q6 & \\
    LAB\_Sodium\_mEq/L & \\
    \_Q6 & \\
    LAB\_Urea Nitrogen\_mg/dL & \\
    \_Q2 & \\
    LAB\_L\_no\_unit & \\
    \_Q2 & \\
    \_1h-2h & \\
    ATC\_antithrombotic agents & \\
    ATC\_4\_A & \\
    ATC\_SUFFIX\_B01 & \\
    ICU\_STAY\_END & Patient is discharged from ICU \\
    \_Q5 & \makecell[l]{Quantile token indicates the length of stay in ICU. \\ Here it is 5, indicating about average} \\
\end{longtable}

\end{document}